# Neuroevolutionary optimization

Eva Volna[1]

[1] Department of Computer Science, University of Ostrava
Ostrava, 70103, Czech Republic

**Abstract**
This paper presents an application of evolutionary search procedures to artificial neural networks. Here, we can distinguish among three kinds of evolution in artificial neural networks, i.e. the evolution of connection weights, of architectures, and of learning rules. We review each kind of evolution in detail and analyse critical issues related to different evolutions. This article concentrates on finding the suitable way of using evolutionary algorithms for optimizing the artificial neural network parameters.

## 1. Introduction

Research on potential interactions between connectionist learning system, i.e. artificial neural networks, and evolutionary search procedures has attracted a lot of attention recently. We can distinguish among three kinds of evolution in artificial neural networks, i.e. the evolution of connection weights, of architectures, and of learning rules. Under neuroevolution we can understand the connection of evolutionary algorithms and artificial neural networks - that is the using of evolutionary algorithm properties in suggestion of artificial neural network architecture and upon work with them.

Evolutionary algorithms are the term for different approaches as of using the models of evolutionary processes, which have nothing common with biology. They try to use the conception of driving forces of organism's evolution for optimization purposes. Evolutionary algorithms refer to a class of population-based stochastic search algorithms that are developed from ideas and principles of natural evolution. They include: evolution strategies, evolutionary programming, genetic algorithms etc. All these models work with random changes of submitted solutions. Optimization is considered here as a synonym for minimization. This is not a problem because going in search the function maximum is equivalent to going in search of function minimum multiplied by -1. One important feature of all these algorithms is their population-based search strategy. Individuals in a population compete and exchange information with each other in order to perform certain tasks. The individual within the evolutionary algorithm is then the problem solution. If a new solution is better, it substitutes the previous one. The choice of the right representation of individuals and their fitness create the essence of the advantageousness of the evolutionary algorithm, which depends on the selection of suitable choice of evolutionary algorithm and its appropriate operators.

An artificial neural network is characterized by its pattern of connections between the neurons (architecture), its method of determining the weights on the connections (adaptation), and its activation function. Neural network architecture can be described as a directed graph in which each neuron $i$ performs a transfer function $f_i$ of the form (1):

$$y_i = f_i\left(\sum_{j=1}^{n} w_{ij} x_j - \theta_i\right) \qquad (1)$$

where $y_i$ is the output of the neuron $i$, $x_j$ is the $j^{th}$ input to neuron $i$ and $w_{ij}$ is the connection weight between neurons $i$ a $j$. $\theta_i$ is the threshold (or bias) of the neuron $i$. Usually, the activation function $f_i$ is nonlinear, such as a sigmoid, or Gaussian function. Learning in artificial neural networks can roughly be divided into supervised, unsupervised, and reinforcement learning. Supervised learning is based on direct comparison between the actual output of an artificial neural network and the desired correct output, also known as the target output. It is often formulated as the minimization of an error function such as the total mean square error between the actual output and the desired output summed over all available data (2):

$$E = \frac{1}{2}\sum_{j=1}^{m}\sum_{i=1}^{n}(y_i - t_i)_j^2 \qquad (2)$$

where $y_i$ is the actual output of the neuron $i$, $t_i$ is the desired correct output of the neuron $i$, $n$ is a number of output neurons, and $m$ is a number of training patters. A gradient descent-based optimization algorithm such as backpropagation can be used to adjust connection weights in the artificial neural network iteratively in order to minimize the error (2). Reinforcement learning is a special case of supervised learning where the exact desired output is unknown. It is based only on the information of whether or not the actual output is correct. Unsupervised learning is solely based on the correlations among input data. The essence of a learning algorithm is *the learning rule*, i.e., a weight-updating rule that determines how connection weights are changed. Examples of popular learning rules include the delta rule, the Hebbian rule, the competitive learning rule, etc. are discussed in numerous publications [1].

## 2. Evolution in artificial neural networks

Evolution has been introduced into artificial neural networks at roughly three different levels: connection weights, architectures, and learning rules. The *evolution of connection weights* introduces an adaptive and global approach to problem solution. The *evolution of architectures* enables artificial neural networks to adapt their topologies to different tasks without human intervention and thus provides an approach to automatic artificial neural network design. The *evolution of learning rules* can be regarded as a process of "learning to learn" in artificial neural networks, where the adaptation of learning rules is achieved through evolution.

2.1 The evolution of connection weights

The evolutionary approach to weight training in artificial neural networks consists of two major phases. The first phase means to decide the representation of connection weights. The second one means the evolutionary process simulated by evolutionary algorithms.

The most convenient representation of connection weights is, from evolutionary algorithm's perspective, *binary* string. In such a representation scheme, each connection weight is represented by a number of bits of a certain length. An artificial neural network is encoded by concatenation of all the connection weights of the network into the chromosome. The order of the concatenation is, however, essentially ignored, although it can affect the performance of evolutionary training, e.g. training time and accuracy. The advantages of the binary representation lie in its simplicity and generality. It is straightforward to apply classical crossover (such as one-point or uniform crossover) and mutation to binary strings. A limitation of binary representation is the representation precision of discrete connection weights. It is still an open question how to optimize the number of bits for each connection weight, the range encoded, and the encoding method used although dynamic techniques could be adopted to alleviate the problem.

To overcome some shortcomings of the binary representation scheme, *real* numbers themselves proposed to represent connection weights directly, i.e. one real number per connection weight. The chromosome is represented by the concatenation of these real numbers, where their order is important. As connection weights are represented by real numbers, each individual in an evolving population is a real vector. Standard search operators dealing with binary strings cannot be applied directly in the real representation scheme. In such circumstances, an important task is to design carefully a set of genetic operators, which are suitable for the real representation as well as artificial neural network's training, in order to improve the speed and accuracy of the evolutionary training. Single real numbers are often changed by average crossover, random mutation or other domain specific genetic operators. It is discussed in [2]. The major aim is to retain useful functional blocks during evolution, i.e., to form and keep useful feature detectors in an artificial neural network.

Evolutionary algorithms are usually based on a global search algorithm, thus can escape from a local minimum, while a gradient descent algorithm can only find a local optimum in a neighborhood of the initial solution. An evolutionary algorithm sets no restriction on types of artificial neural networks being trained as long as a suitable fitness function can be defined properly, thus can deal with a wide range of artificial neural networks: recurrent artificial neural networks, high-order artificial neural networks, fuzzy artificial neural networks etc. An assignment of the most acceptable evolutionary algorithm to a task represents always a big problem, because each search procedure is suitable only for a class of error (fitness) functions with certain types of landscape, the issue of what kind of search procedure is more suitable for which class of error (fitness) function is an important research topic of general interest. The efficiency of evolutionary training can be improved significantly by incorporating a local search procedure into the evolution, i.e., combining evolutionary algorithm's global search ability with local search's ability to fine tune. Evolutionary algorithms can be used to locate a good region in the space and a local search procedure is used to find a near-optimal solution in this region. The obtained results showed that the hybrid GA/BP approach was more efficient than if either the genetic or backpropagation algorithm alone were used, because genetic algorithms are much better at local good initial weights than the random start backpropagation method. Similar work on the evolution of initial weights has also been done on competitive learning neural networks and Kohonen networks [3]. One of the problems faced by evolutionary training of artificial neural networks is the *permutation problem* [4], also known as the *competing convention problem*. It is caused by the many-to-one mapping from the representation (genotype) to the actual artificial neural network (phenotype) since two artificial neural networks that order their hidden neurons differently in their chromosomes will still be equivalent functionally. The permutation problem makes the crossover operator very inefficient and ineffective in producing good offspring. It is generally very difficult to apply crossover operators in evolving connection weights since they tend to destroy feature detectors found during the evolutionary process, because hidden nodes are in essence feature extractors and detectors.

2.2 The evolution of architectures

The architecture of an artificial neural network includes its topological structure, i.e., connectivity, and the transfer function of each neuron in the artificial neural network. Architecture design is crucial in the successful application of artificial neural networks because the architecture has significant impact on a network's information processing capabilities. Up to now, architecture design is still very much a human expert's job. It depends heavily on the expert experience and a tedious trial-and-error process. There is no systematic way to design a near-optimal architecture for a given task automatically. Constructive / destructive algorithms are one of the many efforts made towards the automatic design of artificial neural network architecture. A *constructive algorithm* starts with a minimal network (e.g. network with minimal number of hidden layers, neurons, and connections) and adds new layers, neurons, and connections when necessary during training while a *destructive algorithm* does the opposite, i. e., starts with the maximal network and deletes unnecessary layers, neurons, and connections during training. These methods are susceptible to becoming trapped at local optima, and in addition, they only investigate restricted topological subsets rather than the complete class of network architectures. The design of the optimal artificial neural network architecture can be formulated as a search problem in the architecture space where each point represents some architecture. The performance level of all architectures forms a discrete surface in the space. The optimal architecture design is equivalent to finding the highest point on this surface. There are several characteristics of such a surface, which make the evolutionary algorithms a better candidate for searching the surface than the constructive and destructive algorithms. These characteristics are [5] the following:

- The surface is *infinitely large* since the number of possible neurons and connections is unbounded.
- The surface is *no differentiable* since changes in the number of neurons or connections are discrete and can have a discontinuous effect on artificial neural network's performance.
- The surface is *complex* and *noisy* since the mapping from an architecture to its performance is indirect and dependent on the evaluation method used.
- The surface is *deceptive* since similar architectures may have quite different performance.
- The surface is *multimodal* since different architectures may have similar performance.

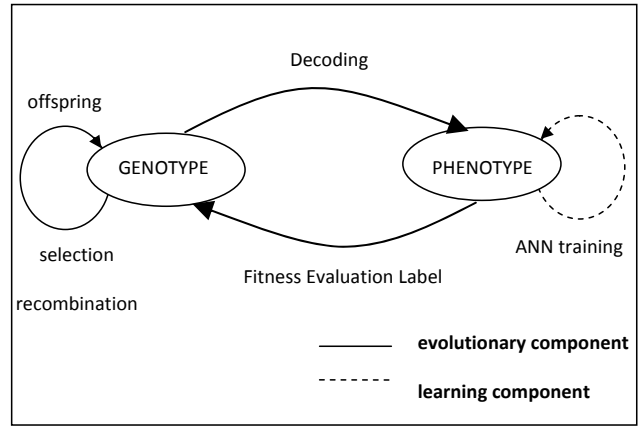

- Fig. 1 Process of evolutionary design of artificial neural networks.

Similar to the evolution of connection weights, two major phases involved in the evolution of architectures are the genotype representation scheme of architectures and the evolutionary algorithm used to evolve artificial neural network architectures. But the problem now is not whether to use a binary representation or a real one, since we only deal with discrete values, a binary representation is required. A key issue here is to decide how much information about architecture should be encoded into a representation. As we apply discrete values, we use usually a binary representation, i.e. matrices or graphs. The evolution of neural network architecture is shown in figure 1. The stochastic optimization algorithms are in principle the only systematic approach to optimization of neural network architecture. Continuous research on evolving neural network architecture has been carried out in the recent years, e.g. [6, 7].

Two different approaches have been taken in the *direct encoding scheme*. In the direct encoding scheme, each connection between neurons is directly specified by its binary representation. It is very suitable for the precise and deterministic search of compact artificial neural network architecture, since a single connection can be added or removed from the artificial neural network easily. One potential problem of the direct encoding scheme is scalability. A large artificial neural network would require a very large matrix and thus increase in the computation time of evolution. In [8] is shown that the length of the genotype is proportional to the complexity of the corresponding phenotype, and the space to be searched by the evolutionary process increases exponentially with the size of the network. Another problem of direct encoding schemes is the impossibility to encode repeated structures (such as network composed of several sub-networks with similar local connectivity) in a compact way. In one-to-one mappings, in fact, elements that are repeated at the level of the phenotype must be repeated at the level of the genotype as well. The next problem to be considered in binary representation is that of variable-length genomes. For example, where two parents have different topologies, it is not obvious how their offspring should be formed. When determining which nodes and connections the offspring should inherit, it would be helpful to know which subnetworks from the parents perform the same functions, and which represent disjoint concepts. Unfortunately, this information is not readily apparent from the two different topologies. Handling a binary representation we strive to take advantages from combining different solutions, while we draw attention to the confrontation between the flexibility of representations and compatibility of genotypes.

In order to reduce the length of the genotype representation of architectures, we can use the *indirect encoding scheme*, where only the most important characteristics of architecture are encoded in the chromosome. The details about each connection in an artificial neural network is either predefined according to prior knowledge or specified by a set of deterministic developmental rules. Many of the indirect neural networks encoding strategies are inspired by the Lindenmayer systems. The typical approach of such encoding is a grammatical encoding [8], where evolutionary algorithms do not develop network architecture directly, but rules of formal grammars, being subsequently used for generating the network topology. The shift from the direct optimization of architectures to the optimization of developmental rules has brought some benefits, such as more compact genotype representation. The rules do not grow with the size of artificial neural networks, since the rule size does not change. The rule is usually described by a recursive equation or a generation rule is similar to a production rule in a production system with a left-hand side and a right-hand side. In [9], a genetic encoding scheme for neural networks based on a cellular duplication and differentiation process was proposed. Genomes are programs written in a specialized graph transformation language called the grammar tree, which is very compact. The genotype-to-phenotype mapping starts with a single cell that undergoes a number of duplication and transformation processes ending up in a complete neural network. In this scheme the genotype is a collection of rules governing the process of cell divisions (a single cell is replaced by two "daughter" cells) and transformations (new connections can be added and the strengths of the connections departing from a cell can be modified). In this model, therefore, connection links are established during the cellular duplication process. This mechanism allows the genotype-to-phenotype process to produce repeated phenotype structures (e.g. repeated neural sub-networks) by reusing the same genetic information, which saves space in genome and it is useful even for keeping the substructures when applying the crossover operator. The literature [10] introduces a new algorithm based on Gene Expression Programming that performs a total network induction using linear chromosomes of fixed length that map into complex neural networks of different sizes and shapes. The total induction of neural networks using gene expression programming requires further modification of the structural organization developed to manipulate numerical constants and domain-specific operators. The indirect encoding scheme is biologically more plausible as well as more practical, from the view point of engineering, than the direct encoding scheme although some fine-tuning algorithms might be necessary to further improve the result of evolution. Other techniques of indirect neural network encoding topology are listened in numerous publications, e.g. [11, 12, 13].

The representation of artificial neural network architectures always plays an important role in the evolutionary design of architectures. There is not a single method, which outperforms others in all aspects. The best choice depends heavily on applications at hand and available prior knowledge. A problem closely related to the representation issue is the design of genetic operators. However, the use of crossover appears to be inconsistent, because crossover works the best when building blocks exist but it is unclear what a building block might be in an artificial neural network since the artificial neural networks are featured with a distributed (knowledge) representation. The knowledge in an artificial neural network is distributed among all the weights in the artificial neural network. Recombining one part of an artificial neural network with another part of another artificial neural network is likely to destroy both artificial neural networks. However, if artificial neural networks do not use a distributed representation but rather a localized one, such as radial basis function networks or nearest-neighbor multilayer perceptrons, crossover might be a very useful operator [14]. In general, artificial neural networks using distributed representation are more compact and have a better generalization capability for most practical problems. As for the evolution of connection weights, thus even here we have to resolve the permutation problem that causes enormous redundancy in the architecture space. Unfortunately, no satisfying technique has been implemented to tackle this problem. An advantage of the evolutionary approach is that the fitness function can be defined easily in such a way that an artificial neural network with some special features is evolved. For example, artificial neural networks with a better generalization can be obtained if testing results, instead of training results, are used in their fitness calculations. A penalty term in the fitness function for complex connectivity can also help improve artificial neural network's generalization ability, besides the cost benefit, by reducing the number of neurons and connections in artificial neural networks.

The discussion on the evolution of architectures so far only deals with the topological structure of architecture. *The transfer function* of each neuron in the architecture has been assumed to be fixed and predefined by human experts yet. The transfer function has been shown to be an important part of artificial neural network architecture and have significant impact on artificial neural network's performance. In principle, transfer functions of different neurons in an artificial neural network can be different and decided automatically by an evolutionary process, instead of assigned by human experts. The decision on how to encode transfer functions in chromosome depends on how much prior knowledge and computation time is available. In general, neurons within a group, like a layer, in an artificial neural network tend to have the same type of transfer function with possible difference in some parameters, while different groups of neurons might have different types of transfer functions. This suggests some kind of indirect encoding method, which lets developmental rules to specify function parameters if the function type can be obtained through evolution, so that more compact chromosomal encoding and faster evolution can be achieved.

2.3 Simultaneous evolution of architectures and connection weights

The evolutionary approaches discussed so far in designing artificial neural network architecture evolve architectures only, without any connection weights. Connection weights have to be learned after a near-optimal architecture is found. This is especially true if one uses the indirect encoding scheme of network architecture. One major problem with the evolution of architectures without connection weights is *noisy fitness evaluation* [15, 16]. In other words, fitness evaluation is very inaccurate and noisy because a phenotype's (i.e., an artificial neural network with a full set of weights) fitness was used to approximate its genotype's (i.e., an artificial neural network without any weight information) fitness. There are two major sources of noise [15]:

- The first source is the random initialization of the weights. Different random initial weights may produce different training results. Hence, the same genotype may have quite different fitness due to different random initial weights used in training.
- The second source is the training algorithm. Different training algorithms may produce different training results even from the same set of initial weights. This is especially true for multimodal error functions.

In order to reduce such noise, architecture usually has to be trained many times from different random initial weights. The average result is then used to estimate the genotype's mean fitness. This method increases the computation time for fitness evaluation dramatically. It is one of the major reasons why only small artificial neural networks were evolved in this way. In essence, the noise is caused by the *one-to-many* mapping from genotypes to phenotypes. It is clear that the evolution of architectures without any weight information has difficulties in evaluating fitness accurately. One way to alleviate this problem is to evolve artificial neural network architectures and connection weights simultaneously [17]. In this case, each individual in a population is a fully specified artificial neural network with complete weight information. Since there is a *one-to-one* mapping between a genotype and its phenotype, fitness evaluation is accurate.

2.4 The evolution of learning rules

An artificial neural network training algorithm may have different performance when applied to different architectures. The design of training algorithms, more fundamentally the learning rules used to adjust connection weights, depends on the type of architectures and learning tasks under investigation. After selecting a training algorithm, there are still algorithm parameters, like the learning rate and momentum in backpropagation algorithms, which have to be specified. For example genetic algorithms are suitable for training artificial neural networks with feedback connections and deep feedforward artificial neural networks (with many hidden layers) while backpropagation is good at training shallow ones. At present, this kind of search for an optimal (near optimal) learning rule can only be done by some experts through their experience and trial-and-error. In fact, what is needed from an artificial neural network is its ability to adjust its learning rule adaptively according to its architecture and the task to be performed. Since evolution is one of the most fundamental forms of adaptation, then said evolution may contribute to the development of appropriate type of the learning rule for given application; for which also the fact may be utilized that the relationship between evolution and learning is extremely complex. Various models have been proposed, but most of them deal with the issue of how learning can guide evolution and the relationship between the evolution of architectures and that of connection weights. Research into the evolution of learning rules is still in its early stages, see e.g. [18, 19]. This research is important not only in providing an automatic way of optimizing learning rules and in modeling the relationship between learning and evolution, but also in modeling the creative process since newly evolved learning rules can deal with a complex and dynamic environment.

The *adaptive adjustment of algorithmic parameters* through evolution could be considered as the first attempt of the evolution of learning rules, e.g. in [20] encoded backpropagation's parameters in chromosomes together with the artificial neural network architecture. The evolution of algorithmic parameters is certainly interesting but it hardly touches the fundamental part of a training algorithm, i.e., its learning rule or weight-updating rule.

Adapting a *learning rule* through evolution is expected to enhance the artificial neural network's adaptivity greatly in a dynamic environment. It is much more difficult to encode dynamic behaviours, like the learning rule, than to encode properties, like the architecture and connection weights, of an artificial neural network. The key issue here is how to encode the dynamic behavior of a learning rule into static chromosomes. Trying to develop a universal representation scheme, which can specify any kind of dynamic behaviors, is clearly impractical, let alone the prohibitive long computation time required searching such a learning rule space. Constraints have to be set on the type of dynamic behaviors, i.e., the basic form of learning rules being evolved in order to reduce the representation complexity and the search space. Two basic assumptions which have often been made on learning rules are [21]: a) weight updating depends only on local information such as the activation of the input neuron, the activation of the output neuron, the current connection weight, etc., and b) the learning rule is the same for all connections in an artificial neural network. A learning rule is assumed to be a linear function of these local variables and their products. That is, a learning rule can be described by the function (3):

$$\Delta w(t) = \sum_{k=1}^{n} \sum_{i_1,i_2,...,i_k=1}^{n} \left( \theta_{i_1,i_2,...,i_k} \prod_{j=1}^{k} x_{i_j}(t-1) \right) \quad (3)$$

where $t$ is time, $\Delta w$ is the weight change, $x_1, x_2,...,x_n$ are local variables, and the $\theta$'s are real-valued coefficients, which will be determined by evolution. In other words, the evolution of learning rules in this case is equivalent to the evolution of real-valued vectors of $\theta$'s. The major aim of the evolution of learning rules is to decide these coefficients. Different $\theta$'s determine different learning rules. Due to a large number of possible terms in (3), which would make evolution very slow and impractical, only a few terms have been used in practice according to some biological or heuristic knowledge [22]. Research related to the evolution of learning rules is also included in [23, 24], although they did not evolve learning rules explicitly. Researchers emphasized the crucial role of the environment in which the evolution occurred.

## 3. Conclusions

*Optimization within informatics* means to seek the answer to the question "*which solution would be the best*" for a problem, in which the quality of each answer may be evaluated via a single value. Although we commonly use the word "*optimum*"; in practice we should obtain the exact global optimum within a huge complex space, which may be considered here with troubles only. Generally, solving the practical tasks, we need sufficient enough approximated (suboptimum) resolution however, above mentioned need not be implicitly a global optimum. Criterion "*sufficient enough*" differs for various types of solved problems. Evolution course usually endeavors to find out a certain task suboptimum solution, instead of exact one.

*Optimization within artificial neural networks* means to seek the optimal combinations of architecture, learning rule and connection weights. Global search procedures such as evolutionary algorithms are usually computationally expensive. It would be better not to employ evolutionary algorithms at all three levels of evolution. It is, however, beneficial to introduce global search at some levels of evolution, especially when there is little prior knowledge available at that level and the performance of the artificial neural network is required to be high, because the trial-and-error and other heuristic methods are very ineffective in such circumstances. Due to different time scales of different levels of evolution, it is generally agreed that global search procedures are more suitable for the evolution of architectures and that of learning rules on slow time scales, which tends to explore the search space in coarse grain (locating optimal regions), while local search procedures are more suitable for the evolution of connection weights on the fast time scale, which tends to exploit the optimal region in fine grain (finding an optimal solution). Such designed artificial neural networks have been shown to be quite competitive in terms of the quality of solutions found and the computational cost. With the increasing power of parallel computers, the evolution of large artificial neural networks becomes feasible. Not only can such evolution discover possible new artificial neural network architectures and learning rules, but it also offers a way to model the creative process as a result of artificial neural network's adaptation to a dynamic environment.

**Eva Volna** She graduated at the Slovak Technical University in Bratislava and defended PhD. thesis with title "Modular Neural Networks". She has been working as a lecturer at the Department of Computer Science, University of Ostrava (Czech Republic) from 1992. Her interests include artificial intelligence, artificial neural networks, evolutionary algorithms, and cognitive science. She is author more than 50 scientific publications.